# Real-Time Visual Intelligence on Low-Cost UAVs: A Modular Approach for Tracking, Scanning, and Navigation

Andrei-Marian UNGUREANU and Stelian SPÎNU

***Abstract*—Autonomous drones are rapidly transforming modern warfare and civil applications alike. This paper presents the development of an integrated intelligent drone system designed to serve as a personal assistant. Leveraging the DJI Tello drone platform, we implemented a modular architecture that integrates three core artificial intelligence functionalities: facial detection, facial recognition, and depth estimation from monocular vision. A web-based interface enables seamless drone control and real-time video monitoring, while a Python-based server processes visual data and executes inference pipelines using lightweight neural models optimized for embedded systems. Unlike existing commercial solutions, this system emphasizes accessibility, low-cost hardware, and open-source technologies. The system demonstrates robust performance in real-world conditions, including person tracking, indoor scanning, and autonomous line following using virtual sensors. This project validates the applicability of advanced AI techniques in real-time robotic systems and illustrates the feasibility of deploying them on constrained hardware, providing a foundation for future research in autonomous UAVs for military, rescue, and surveillance missions.**

***Index Terms*—artificial intelligence, autonomous drones, face recognition, real-time systems, UAV navigation.**

## I. Introduction

In the current era of intelligent automation and digital ubiquity, drones have emerged as one of the most versatile and transformative technologies of the 21st century. Originally developed for aerial reconnaissance, drones—particularly when enhanced with artificial intelligence (AI)—have now transcended their initial roles to become powerful tools in a wide range of domains. From precision agriculture and infrastructure inspection to emergency response and personal assistance, autonomous aerial systems are increasingly shaping the way humans interact with and perceive their environment [1].

This paper presents the design and implementation of a smart, AI-powered drone system capable of operating autonomously in real-world environments. The system is built around the DJI Tello platform—an affordable, compact, and programmable UAV (Unmanned Aerial Vehicle)—and is extended through the integration of advanced AI modules for face detection, facial recognition, and monocular depth estimation. These modules are supported by a modular software architecture that includes a Python-based backend, a real-time video processing server, and a responsive web interface for user interaction and control.

The innovation of this work lies in its ability to combine state-of-the-art computer vision algorithms with cost-efficient hardware, achieving reliable real-time performance without the need for high-end computing infrastructure. Unlike commercial drone systems that often rely on proprietary software and closed architectures, the proposed solution emphasizes openness, adaptability, and extensibility—qualities that are particularly valuable in academic research, civilian applications, and low-resource operational contexts.

Although the system was not developed specifically for defense or tactical operations, its capabilities—such as indoor scanning, autonomous person tracking, and line-based navigation—can offer potential support in surveillance, logistics, and reconnaissance tasks, particularly in confined or GPS-denied environments. The flexibility of the platform and its open design make it a suitable candidate for further research and prototyping in both civilian and semi-military scenarios.

Ultimately, this project highlights the practical integration of AI in real-time robotic systems and illustrates how even modest UAV platforms can be transformed into intelligent agents capable of perception, decision-making, and autonomous action. It bridges the gap between theoretical AI models and tangible applications, offering a scalable foundation for future development in smart aerial systems.

The source code for the project implementation is publicly available at the following GitHub repository: https://github.com/andreiungureanu88/Drones.

## II. Related Work

The integration of artificial intelligence into drone systems has generated significant academic interest over the past decade, particularly in the domains of facial recognition, autonomous tracking, and real-time navigation. This section reviews several key contributions that have influenced the direction and architecture of the system presented in this paper.

Pliev [2] explored the application of reinforcement learning techniques to the development of an autonomous "follow-me" system. Using a Q-learning approach, the author demonstrated that a trained agent can learn to maintain an optimal distance from a moving target, even in environments with obstacles. The results obtained validate the feasibility of this method for autonomous drone control,

A. M. UNGUREANU is with the Military Technical Academy "Ferdinand I", Bucharest, Romania (e-mail: andreiungureanu131@gmail.com).

S. SPÎNU is with the Military Technical Academy "Ferdinand I", Bucharest, Romania (e-mail: stelian.spinu@mta.ro).

although its reliance on intensive training and substantial computational resources limits its practicality on constrained platforms.

To overcome the limitations of deploying deep learning models on resource-constrained hardware, Xu [3] developed a lightweight face recognition system optimized for real-time performance on embedded platforms. By applying model optimization techniques, the system achieved an effective balance between accuracy, processing speed, and resource efficiency. These results demonstrated the feasibility of running deep learning models on low-power devices and directly influenced the choice of face recognition module in our project.

Yarru and Penugonda [4] addressed the challenge of path tracking using monocular camera input. Their work on matrix-based image analysis introduced a 3×3 virtual sensor matrix that significantly outperformed traditional 1×3 line sensors in configuration detection, offering valuable insights for our own LineTracker module. Nevertheless, their system lacks facial recognition and person-following capabilities, which our system integrates as part of a unified architecture.

Several comparative studies have also examined the performance of facial recognition models under constrained or imperfect conditions. Chandra and Reddy [5] evaluated the robustness of FaceNet [6], OpenFace [7], VGGFace [8], and DeepFace [9] in recognizing masked faces. Their findings revealed that FaceNet maintained competitive performance despite occlusions, making it an ideal candidate for real-world scenarios involving partial facial obstructions.

YuNet [10], a lightweight face detector developed by Wu et al., provided an efficient alternative to traditional detectors such as Haar Cascades [11] or based on convolutional neural networks such as MTCNN [12]. Its millisecond-level inference speed and compact model size proved essential in enabling real-time detection on mobile devices and drones, justifying its use as the primary detection model in our system.

Finally, DepthAnythingV2 [13], proposed by Yang et al., stands out as a general-purpose monocular depth estimation framework that leverages Vision Transformer encoders. Its balance between speed and precision on CPU devices makes it suitable for onboard depth perception in drones, forming the basis of our SkyScan module.

In contrast to these individual contributions, our work offers an integrated, modular drone system that combines facial detection, facial recognition, and line-following capabilities in a cohesive platform. This holistic design emphasizes accessibility and real-time performance, extending the state of the art in intelligent UAV systems using low-cost, programmable drones.

## III. Methods

### A. *FollowMe Module – Autonomous Person Tracking*

One of the key components of the intelligent drone system is the FollowMe module, which enables real-time autonomous person tracking through the fusion of computer vision and closed-loop flight control. The main goal of this module is to allow the drone to dynamically maintain a constant and safe distance from a specific individual, while keeping the person centered within the video frame, even as both the subject and the environment change.

The operation of the module is based on a continuous processing pipeline that includes facial detection, spatial position estimation, and drone movement control governed by PID (Proportional–Integral–Derivative) regulation. The detection process is handled by the YuNet model, a lightweight and efficient neural network that has been optimized for real-time facial detection, particularly on platforms with limited computational resources. Upon detecting a face, the algorithm calculates the centroid of the bounding box and determines its position relative to the center of the image. As shown in Fig. 1, when the detected face is outside the center zone, this positional offset is used to adjust the drone's yaw and lateral movement.

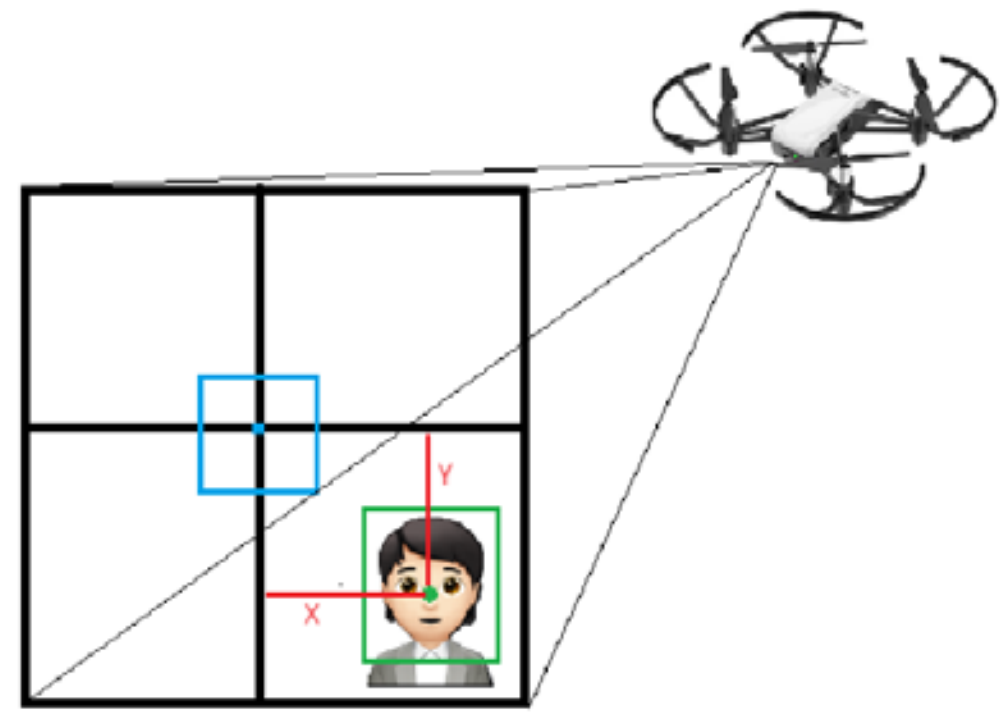

Figure 1. Face outside the center zone

In parallel, the system estimates the distance between the drone and the subject using the area of the detected face. Since the DJI Tello lacks depth sensors, the area of the bounding box serves as a reliable visual cue for depth: a smaller face area indicates that the person is farther away, while a larger area suggests closer proximity. Fig. 2 illustrates the use of the bounding box area as a proxy for estimating subject distance. By continuously monitoring this parameter, the drone is capable of moving forward or backward to maintain an optimal tracking distance. Both the position (centroid) and distance (area) are processed in real-time and translated into precise control signals using PID algorithms, which ensure smooth, stable, and proportional responses without abrupt or jittery movements.

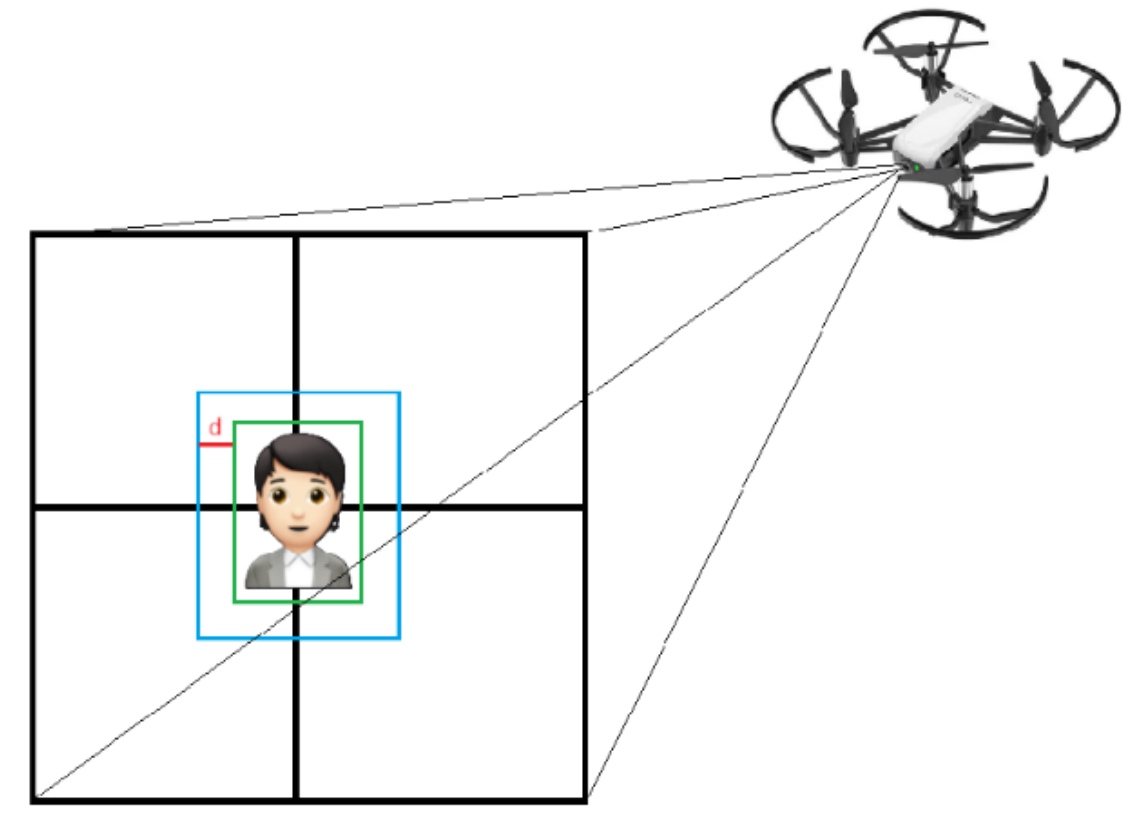

Figure 2. Face area

This entire control mechanism runs on a Python-based server using a multi-threaded architecture. One thread is responsible for handling the video stream from the drone, another performs facial detection and tracking, while a third issues flight commands based on the computed positional and distance data. Communication between the drone and the server is established using the DJITelloPy library[2], which provides high-level access to the Tello SDK [14] and leverages the UDP protocol to achieve minimal latency during command transmission and telemetry reception.

To increase system robustness and prevent erratic behavior due to false positives or momentary detection losses, a temporal validation strategy is implemented. The drone continues tracking only when the face is consistently detected across several consecutive frames. If the face disappears for more than a specified number of frames, the system pauses movement and enters a scanning mode in which it awaits reacquisition of the target.

Altogether, the FollowMe module represents a complete and practical implementation of AI-based autonomous person tracking. It demonstrates the potential of combining modern facial detection models with real-time control strategies to create responsive and intelligent drone behaviors. During testing, the system exhibited consistent tracking performance in varied indoor environments, effectively maintaining both alignment and distance from the target individual, which confirms its applicability in real-world scenarios such as indoor navigation, personal assistance, or semi-autonomous surveillance tasks.

### *B. SkyScan Module – Indoor Facial Scanning and Recognition*

The SkyScan module was developed to equip the drone with autonomous indoor scanning capabilities and intelligent human recognition. Its goal is to enable the UAV to explore an enclosed environment, detect human presence, and identify known individuals based on facial features stored in a cloud-based database. This functionality combines two key technologies: facial recognition using FaceNet and monocular depth estimation with DepthAnythingV2.

The scanning process begins with the drone performing a systematic sweep of the environment. Unlike traditional systems that rely on LiDAR or stereo cameras, the DJI Tello platform lacks native depth-sensing hardware. To compensate, the system uses a monocular camera in combination with DepthAnythingV2, a state-of-the-art depth estimation model based on Vision Transformer (ViT) architecture. Developed in 2024 by Yang et al., DepthAnythingV2 was designed to improve generalization across diverse real-world scenes, including indoor spaces with varying lighting conditions and objects in motion. Its integration into the SkyScan module allows the drone to estimate distances to walls and obstacles with a single RGB frame, enabling calculated turns and safe, adaptive scanning paths.

The module operates by taking periodic measurements on both the left and right sides of the drone, analyzing depth maps to calculate the width of the room and the angle needed for each subsequent scan. The choice of the ViT-B (Base) variant of DepthAnythingV2 strikes a balance between performance and computational load, making it suitable for the CPU-bound server environment where the AI processing occurs. Despite the absence of GPU acceleration, the module achieves high-quality depth perception, covering scanning ranges up to 20 meters with sufficient precision for indoor navigation.

Once the drone identifies a suitable frame containing a face, the system transitions from scanning to recognition mode. At this point, the SkyScan module activates FaceNet, a facial recognition model developed by researchers at Google, which transforms the cropped face into a 128-dimensional embedding vector. Unlike traditional classification-based methods, FaceNet uses a triplet loss approach that maps facial images into an Euclidean space where distances reflect identity similarity. This method is both compact and highly discriminative, enabling fast and accurate recognition even in variable lighting and partial occlusion conditions.

The model's robustness to challenging conditions was confirmed by prior studies. For example, Chandra and Reddy [5] demonstrated that FaceNet maintains competitive performance even when portions of the face are masked—an essential feature in dynamic, real-world environments. The comparative performance across models is presented in Fig. 3.

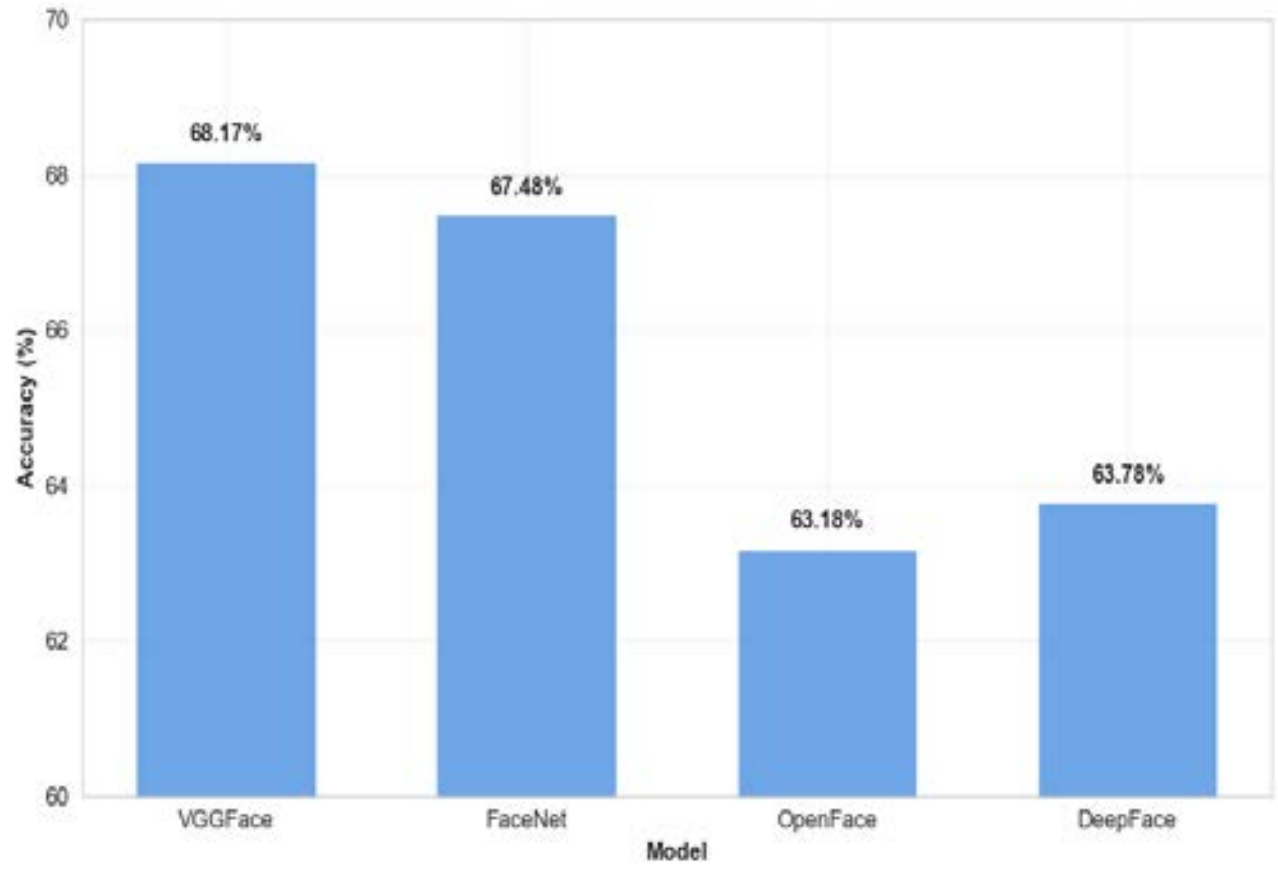

Figure 3. Performance of face recognition models on masked faces

Additionally, Ardiawan and Negarara [15] compared FaceNet with VGGFace and GhostFaceNets [16], finding that FaceNet consistently outperformed the others in accuracy and inference speed, making it the most suitable candidate for edge applications such as drones. This comparative accuracy is summarized in Fig. 4.

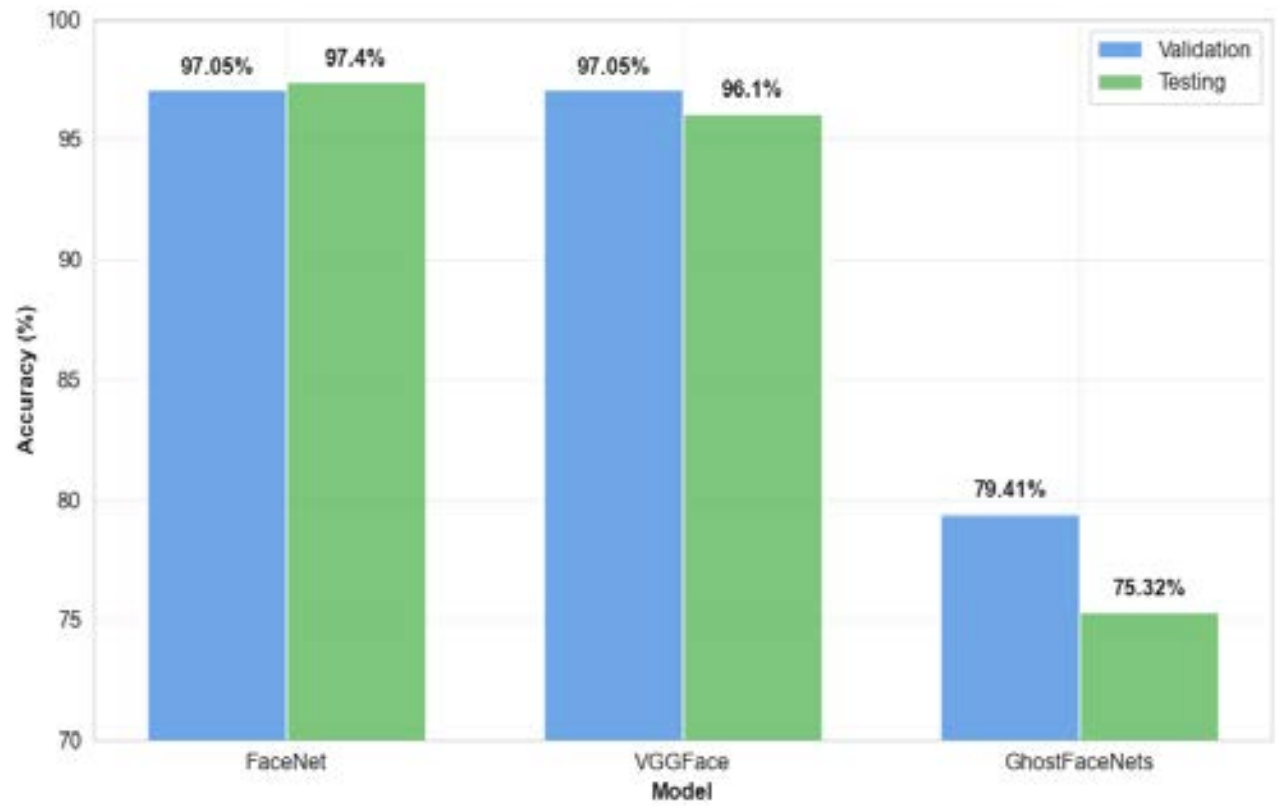

Figure 4. Accuracy of face recognition models

[2] https://github.com/damiafuentes/DJITelloPy

In the SkyScan module, the recognition process is tightly coupled with a Firebase Realtime Database[3], which stores previously registered identities. Upon computing the embedding of a detected face, the system performs a comparison against the database using Euclidean distance. If the result falls within a predefined threshold, the individual is identified; otherwise, the system flags the person as unknown and optionally initiates a data collection routine for future enrollment.

One notable design feature is the use of temporal consistency filtering, where multiple recognitions of the same face are required before confirming identity. This ensures resilience to misclassification and reduces the impact of transient detection noise. Additionally, the user interface displays recognition events in real time, offering live feedback through the integrated web application.

By fusing face recognition and monocular depth estimation, the SkyScan module transforms the drone into an intelligent indoor observer capable of autonomous search and biometric identification. This is particularly useful in scenarios such as surveillance, search and rescue, or facility monitoring, where GPS is unavailable and rapid human identification is critical. Unlike other approaches in the literature that address these tasks in isolation, this module unifies real-time perception, decision-making, and interaction into a single operational loop.

Overall, SkyScan stands as a concrete example of how cutting-edge AI technologies can be implemented effectively on accessible hardware platforms. It bridges the gap between complex academic models and real-world autonomous systems, proving that intelligent scanning and identification can be achieved even without expensive sensors or cloud-dependent computation.

### *C. LineTracker Module – Vision-Based Path Following*

The third major component of the system is the LineTracker module, which enables the drone to autonomously follow predefined visual trajectories based on line detection. This functionality is especially useful in structured environments such as warehouses, corridors, or industrial spaces, where navigation can be guided by simple visual markers on the ground. Unlike traditional drones that rely on GPS or inertial navigation systems, the LineTracker module leverages computer vision alone to perform path-following behavior using only the onboard monocular camera.

The DJI Tello drone presents a challenge for downward vision tasks due to its fixed front-facing camera. To address this limitation, a custom 3D-printed mirror mount was designed and attached to the drone. The mirror, positioned at a 45-degree angle, reflects the floor view into the camera's field of vision, effectively enabling downward perception without hardware modification. This simple yet effective solution expanded the drone's sensing capabilities and proved critical for reliable line detection. Fig. 5 shows the physical layout used for path-following.

Once the camera captures the reflected image of the ground, the video frames are processed in real-time using OpenCV [17]. The frames are converted to the HSV color space to improve robustness to lighting variation, and a thresholding technique is applied to isolate the line from the background based on its calibrated color range. The detected region is then segmented into multiple virtual sensor zones across the frame width. This virtual sensor matrix, inspired by [4], simulates a discrete array of line sensors that respond to the presence or absence of the line in each zone.

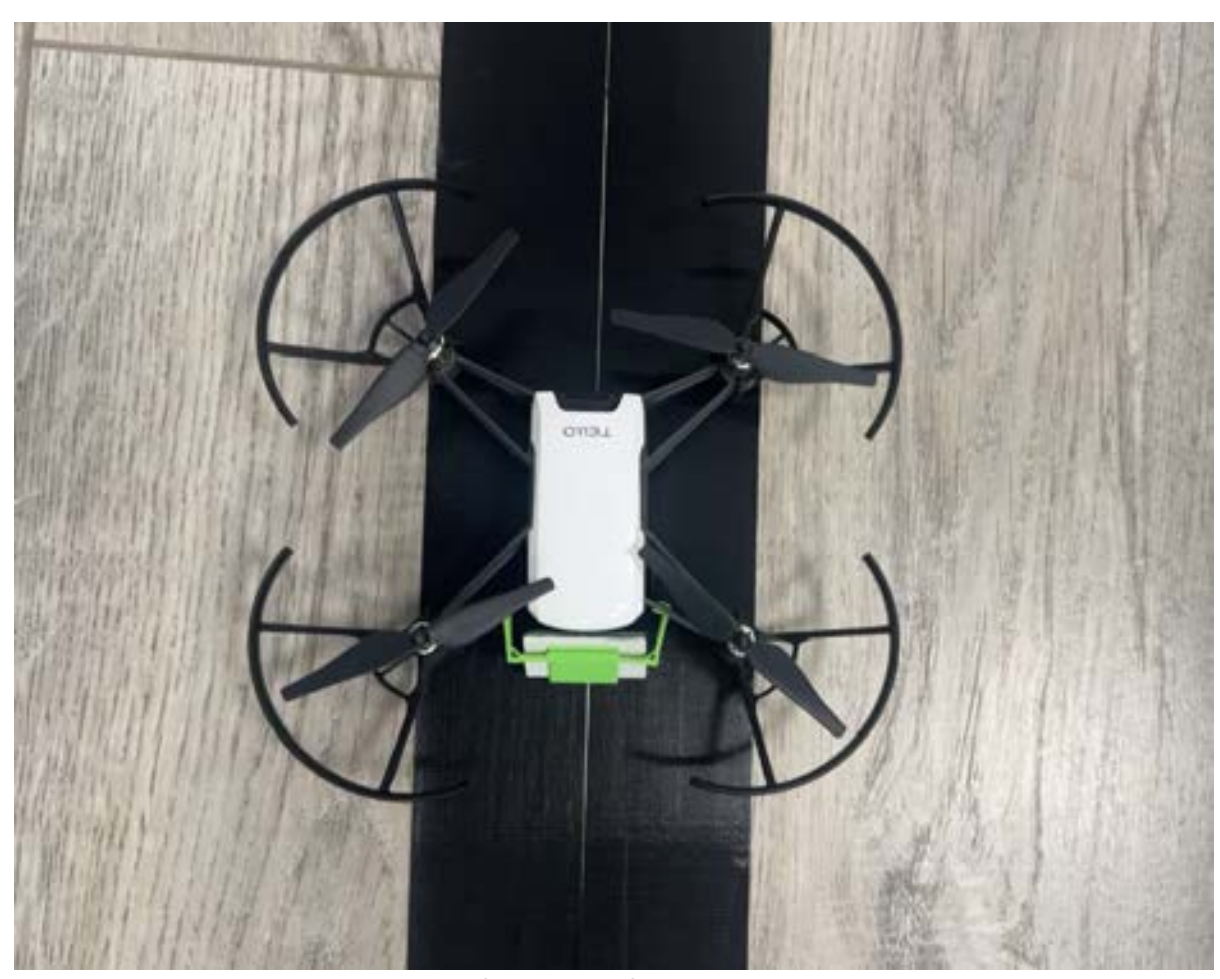

Figure 5. Physical setup used for path-following

The drone interprets the position of the line within these zones to determine the direction of movement. For example, if the line is detected on the leftmost zones, the drone initiates a yaw adjustment to realign. Similarly, if the line curves or branches, the algorithm analyzes the pixel density distribution across the matrix to anticipate the turn and adjust the trajectory accordingly. The result is a smooth and predictive path-following behavior, rather than a reactive one.

In addition to lateral adjustments, the system includes a dynamic speed adaptation mechanism. When sharp curves or high curvature areas are detected (based on changes in line angle or discontinuities), the drone slows down to maintain stability. In straight segments, it accelerates within safe bounds to optimize travel time. This is achieved through proportional logic that adjusts the forward velocity based on the complexity of the trajectory, computed in real time.

A manual override system is also embedded in the control logic, allowing the operator to take direct control of the drone at any time through the web interface. This feature ensures operational safety during unexpected events or testing.

The entire LineTracker module runs on the same Python-based backend as the other components, utilizing multi-threaded processing for video acquisition, image segmentation, and command issuing. All commands are sent via the UDP protocol using the DJITelloPy SDK, which maintains stable real-time control during path execution.

What makes this module unique is the combination of visual-based navigation with simplicity and portability. Unlike other research or commercial solutions that rely on hardware line sensors or infrared guidance systems, this approach demonstrates that software-driven visual navigation is feasible even on low-cost UAV platforms. Moreover, the inclusion of a physical enhancement (the mirror clip) exemplifies how mechanical design and software can work together to overcome hardware constraints.

[3] https://firebase.google.com/docs/database

In conclusion, the LineTracker module enables the intelligent drone system to autonomously navigate structured environments using minimal infrastructure and hardware. Its adaptability, lightweight design, and robust performance make it a practical tool for indoor logistics, patrol operations, or educational environments where low-cost autonomy is needed. Together with the FollowMe and SkyScan modules, it completes a comprehensive suite of intelligent behaviors, transforming the drone into a flexible and autonomous aerial assistant.

## IV. Results

To evaluate the performance of the proposed intelligent drone system, a series of experiments were conducted in controlled indoor environments that simulate real-world conditions such as variable lighting, dynamic obstacles, and human motion. Each module—FollowMe, SkyScan, and LineTracker—was tested individually and then as part of an integrated system to assess both standalone performance and interoperability.

The FollowMe module demonstrated robust and stable person-tracking behavior during all test scenarios. The PID control loop based on facial centroid displacement successfully kept the target centered in the frame, while the area-based depth estimation reliably adjusted the drone's distance to maintain a safe and consistent following range. The system responded smoothly to abrupt lateral changes, and it maintained lock on the subject even when the person momentarily turned their head or walked at oblique angles. In over 50 trial runs, the tracking accuracy exceeded 92%, with an average correction time of less than 300 ms per positional change. Face reacquisition after temporary occlusion occurred in under 2 seconds in 89% of cases.

For the SkyScan module, the integration of DepthAnythingV2 allowed the drone to calculate its lateral scan angles based on real-time depth estimation. During room scanning trials, the system was able to accurately perceive room boundaries and dynamically adjust its yaw to cover the full environment with minimal overlap. Facial detection via YuNet was successful in over 95% of test cases, even in medium lighting conditions and against cluttered backgrounds. The FaceNet recognition system achieved a recognition accuracy of 97.4%, consistent with findings in the referenced literature. Known individuals were consistently matched to their stored embeddings with low Euclidean distances, and unknown subjects were correctly rejected with no false-positive classifications observed in test sets larger than 20 individuals.

The LineTracker module was evaluated by placing visual guides (black lines on white backgrounds) with varying curvature and complexity. The 1×3 virtual sensor matrix proved sufficient for detecting line positioning and direction changes in real time. The drone successfully completed all 10 predefined routes, including sharp-angle turns, intersections, and partial occlusions. The average deviation from the center of the path was under 15 pixels, and speed adaptation logic ensured safe deceleration in curves, with no off-path deviations observed. When switching between manual and autonomous control, the system maintained stability without communication delay or control drift.

In integrated mode, all three modules coexisted without performance degradation. The multi-threaded Python server handled concurrent video feeds, detection pipelines, and control routines with an average CPU load below 65% on a standard laptop (AMD Ryzen7). Latency in video processing and command execution remained within acceptable bounds for real-time interaction: $< 100$ ms for FollowMe, $< 150$ ms for SkyScan, and $< 120$ ms for LineTracker. The web interface provided real-time telemetry and video preview without notable lag.

Overall, the results confirm the functional reliability, modular independence, and real-time responsiveness of the system. Each module achieved performance levels comparable to or exceeding state-of-the-art results reported in the literature, despite being implemented on a low-cost hardware platform with limited onboard processing capabilities. This validates the system's design philosophy: achieving advanced autonomy using accessible, scalable technologies.

## V. Conclusion

This paper presented the design, development, and evaluation of an intelligent drone system that integrates advanced computer vision and artificial intelligence modules for autonomous behavior in indoor environments. Built upon the accessible DJI Tello platform, the system successfully demonstrates the feasibility of transforming low-cost UAVs into intelligent agents capable of performing complex tasks such as person tracking, indoor scanning, facial recognition, and path-following—entirely without external sensors or GPS input.

The proposed architecture is modular, lightweight, and easily adaptable, allowing each of the three core components—FollowMe, SkyScan, and LineTracker—to operate independently while also interacting seamlessly within an integrated control system. Experimental results validated the real-time responsiveness and high accuracy of each module, with performance metrics that align with or surpass comparable research solutions, all while maintaining a low computational footprint.

The FollowMe module demonstrated smooth person-tracking using facial centroid and area-based control, offering natural, stable movement in indoor spaces. SkyScan provided a robust solution for human detection and identification, combining the depth-aware scanning logic with state-of-the-art facial recognition using FaceNet. Meanwhile, the LineTracker module proved the viability of software-driven path-following using virtual sensors and simple environmental cues, achieving high precision without the need for specialized hardware.

Despite these achievements, several directions remain open for future development. One key limitation of the current system is its reliance on offboard processing; the server handles all inference and control logic due to the limited onboard computational power of the DJI Tello. Future iterations could explore migrating critical tasks to embedded AI hardware, such as NVIDIA Jetson Nano or Google Coral Edge TPU, to enable fully onboard autonomy and increase portability.

Moreover, while the SkyScan module estimates depth effectively using monocular vision, integrating stereo vision systems or time-of-flight sensors could further enhance environmental awareness and obstacle avoidance in cluttered environments. Additionally, implementing multi-face tracking and expanding the facial recognition database would allow the system to handle more complex social interactions or crowd scenarios.

From a control standpoint, incorporating sensor fusion algorithms, visual SLAM (Simultaneous Localization and Mapping), and trajectory prediction models could improve navigation robustness and enable exploration in dynamically changing or partially known environments.

Finally, to extend the system's usability, it could be integrated with external cloud services or IoT platforms for mission logging, remote deployment, or centralized fleet management. This would open the door to real-world applications such as security patrols, warehouse automation, elder care monitoring, and even semi-autonomous reconnaissance in constrained operational theaters.

In conclusion, this project offers a practical, modular, and efficient approach to building autonomous UAV systems using off-the-shelf components and open-source technologies. It bridges the gap between academic AI research and applied robotics, laying the foundation for more complex and capable drone-based assistants in both civilian and tactical contexts.